\documentclass{article} 
\PassOptionsToPackage{dvipsnames}{xcolor}
\usepackage{iclr2027_conference,times}

\usepackage{microtype}
\usepackage{hyperref}
\usepackage{xurl}
\usepackage{booktabs}
\usepackage{graphicx}
\graphicspath{{latex/figure/}}
\usepackage{subcaption}
\usepackage{amsmath}
\usepackage{amssymb}
\usepackage{mathtools}
\usepackage{amsthm}
\usepackage{multirow}
\usepackage[dvipsnames]{xcolor}
\usepackage{CJKutf8}
\usepackage{textcomp}
\DeclareUnicodeCharacter{2423}{\textvisiblespace}

\definecolor{darkblue}{rgb}{0, 0, 0.5}
\hypersetup{colorlinks=true, citecolor=darkblue, linkcolor=darkblue, urlcolor=darkblue}

\newcommand{\visiblespace}[0]{␣}
\newcommand{\toksep}{\hspace{0.15em}\textbar\hspace{0.15em}}

\title{LiteToken: Removing Intermediate Merge Residues From BPE Tokenizers}

\author{Yike Sun\quad Haotong Yang\textsuperscript{*}\quad Zhouchen Lin\textsuperscript{*}\quad Muhan Zhang\textsuperscript{*}\\
  \normalfont Peking University, Beijing, China\\
  \small\normalfont\textsuperscript{*}Correspondence: \texttt{\{haotongyang,zlin,muhan\}@pku.edu.cn}}
\hypersetup{pdfauthor={Yike Sun, Haotong Yang, Zhouchen Lin, Muhan Zhang}, pdftitle={LiteToken: Removing Intermediate Merge Residues From BPE Tokenizers}}

\iclrfinalcopy
\begin{document}

\maketitle
\lhead{Preprint}

\begin{abstract}
    Tokenization is fundamental to how language models represent and process text, yet the behavior of widely used BPE tokenizers has received far less study than model architectures and training. In this paper, we investigate intermediate merge residues in BPE vocabularies: tokens that are frequent during merge learning so they're retained in the final vocabulary, but are mostly further merged and rarely emitted when tokenizing the corpus during tokenizer usage. Such low-frequency tokens not only waste vocabulary capacity but also increase vulnerability to adversarial or atypical inputs. We present a systematic empirical characterization of this phenomenon across commonly used tokenizers and introduce LiteToken, a simple method for removing residue tokens. Because the affected tokens are rarely used, pretrained models can often accommodate the modified tokenizer without additional fine-tuning. Experiments show that LiteToken reduces token fragmentation, reduces parameters, and improves robustness to noisy or misspelled inputs, while preserving overall performance.
\end{abstract}

\section{Introduction}
\label{sec:intro}

Although the parameter count, training data, and training compute of large language models (LLMs) have been growing dramatically in recent years, the very first component of LLMs, i.e., the tokenizer, has received less attention than model architectures and training. For most contemporary models, including the most powerful closed-source systems~\citep{gpt4o, geminiteam2025geminifamilyhighlycapable}, tokenization largely preserves long-standing design choices, most notably byte-level Byte Pair Encoding (BPE)~\citep{Gage1994bpe, sennrich-etal-2016-neural,wang2020bbpe}. The impact of tokenizer design on downstream performance remains under-explored, and many tokenization algorithms have been developed and evaluated primarily in regimes with smaller models, datasets, and vocabulary sizes~\citep{schuster2012japanese,kudo-2018-subword,provilkov-etal-2020-bpe}.

\begin{figure}
    \centering
    \includegraphics[width=0.6\linewidth]{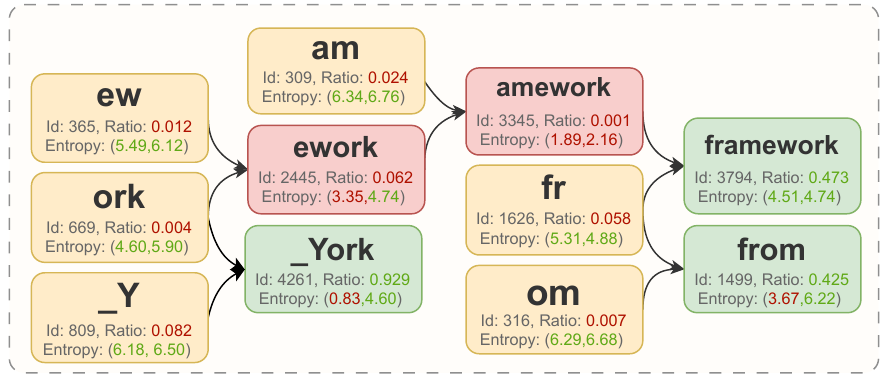}
    \caption{\textbf{An example of merging tree in Qwen tokenizer.} \textcolor{Green}{Green}: Tokens with high FI ratio, high frequency in final tokenized corpus; \textcolor{Goldenrod}{Yellow}: Tokens with low FI ratio but high entropy score, which are meaningful partial tokens. \textcolor{Red}{Red}: Tokens with low FI ratio and low entropy score (fixed combinations), i.e., intermediate merge residues, which should be removed from the vocabulary. Thresholds: 0.25 for the FI ratio and 4.0 for the entropy score. FI ratio represents the ratio between final occurrences of the token in the entire corpus and the total number of times this token appeared during the BPE training merge process, which is formally defined in Section 3.2.1.}
    \label{fig:example_tree}
\end{figure}

However,
growing evidence suggests that improperly designed tokenizers can induce unexpected and undesirable model behaviors, increasing adversarial vulnerability and even triggering harmful generations~\citep{the_art_of_prompt, phan2024understandingmitigatingtokenizationbias,li2024glitchtokenslargelanguage, land2024fishingmagikarpautomaticallydetecting}. 
In this work, we study a commonly overlooked issue in modern tokenization practice that we term \textit{``intermediate merge residues''} (also referred to as \textit{``scaffold tokens''}~\citep{lian2024scaffoldbpeenhancingbytepair})~\citep{chizhov2024bpegetspickyefficient}. Concretely, BPE training begins with an initial vocabulary, counts the frequency of adjacent token pairs in the corpus, and repeatedly merges the most frequent pair to form a new token. During this process, many tokens that are highly frequent \textit{at that moment they are created}, but are later repeatedly subsumed by subsequent merges, ultimately becoming infrequent in the final vocabulary (red tokens in Figure~\ref{fig:example_tree}). 
This phenomenon becomes particularly pronounced for large vocabularies, where longer portions of the merge trajectory are retained in the vocabulary. Empirically, we find that in some widely used LLM tokenizers, about 10\% of tokens can be identified as intermediate merge residues. 


We analyze the impact of intermediate merge residues along three dimensions:
1)~\textbf{Token waste}: Indispensable part in the token embeddings and the decoding (output) layers along with the corresponding decoding computation are not effectively utilized, constituting a tangible waste of capacity and compute, especially for the small models ($\leq$8B parameters)~\citep{lu-etal-2025-demystifying}. As shown in Table~\ref{tab:waste}, for these smaller models, the intermediate merging residues cause about 1\% -3\% parameters and flops waste.
Moreover, it may pose a latent bottleneck for merging research directions such as \textit{multi-token prediction} (MTP)~\citep{gloeckle2024betterfasterlarge,deepseekv3} and \textit{$N$-gram}–based methods~\citep{huang2025overtokenized,cheng2026conditionalmemoryscalablelookup} where token-level \textit{``redundancy''} and \textit{``noise''} will be exponentially amplified.
2)~\textbf{Performance degradation}: These tokens are systematically low-probability under the training corpus distribution and therefore under-trained. As a result, models are more prone to errors when encountering them, particularly in atypical or out-of-distribution inputs.
3)~\textbf{Vulnerability}: Because residue tokens are both infrequent and often associated with noisier text patterns, they can disproportionately harm generation quality and reduce robustness under adversarial perturbations, including jailbreak-style attacks.

\begin{table}[t]
    \centering
    \caption{\textbf{Param and Flops waste with 10\% useless token.} Estimated with the Qwen family's configuration. For the flops, we estimate with input length 256 and full precision. ``-first represents a full forward passing (without KV cache); ``-cache'' represents one new token generation with KV cache.}
    \label{tab:waste}
    \small
    \begin{tabular}{lcccccc}
    \toprule
        Model & 0.6B & 1.7B & 4B & 8B &14B &32B \\ \midrule
        Tied embedding & T & T & T & F & F & F \\
        \#Param (\%) & 2.6 & 1.8 & 1.0& 1.6 & 1.1 & 0.5 \\
        Flops-first (\%) &3.0 & 1.8 & 1.0 & 0.8 & 0.6 & 0.3 \\
        Flops-cache (\%)& 3.2 & 1.9 & 1.1 & 0.8 & 0.6 & 0.3 \\ \bottomrule
    \end{tabular}
\end{table}

To address this, we propose \textbf{LiteToken}, a lightweight algorithm to identify and remove these \textit{``intermediate merge residues''} from an existing BPE tokenizer. Our method first conducts a corpus-driven statistical analysis to detect candidate \textit{``intermediate merge residues''} (red tokens in Figure~\ref{fig:example_tree}), followed by further filtering to retain linguistically meaningful units like some common word roots, prefixes, or suffixes (yellow tokens in Figure~\ref{fig:example_tree}). Then we design a re-merging algorithm that ensures the pruned vocabulary can still tokenize the original corpus effectively, improving token efficiency while better aligning the resulting segmentation with linguistic structure.

Thanks to the low-probability of the pruned tokens, we find that the lite tokenizer can be directly applied to the original model \textit{without any finetuning}, enabling a \textit{plug-and-play} deployment with negligible cost. Empirically, the pruned tokenizer not only reduces model parameters and computational overhead, but also improves robustness to perturbations such as typographical errors and adversarial attacks. 




\section{Related Work}
\label{sec:related_work}
\subsection{Tokenizer learning algorithm}

Modern LLMs typically use sub-word tokenization to balance vocabulary coverage with a manageable vocabulary size. Classic merge-based methods include BPE~\citep{Gage1994bpe}, which repeatedly merges the most frequent adjacent symbol pairs, and WordPiece~\citep{schuster2012japanese}, widely used in BERT-like models~\citep{devlin-etal-2019-bert} with a likelihood-inspired objective and longest-match decoding. Byte-level BPE (BBPE) performs merges over bytes for full character coverage. In contrast, Unigram LM~\citep{kudo-2018-subword} starts from a large seed vocabulary and selects a subset to maximize likelihood, optionally using sampling for regularization. Systems-oriented implementations such as TikToken~\citep{TikToken}, SentencePiece~\citep{kudo-richardson-2018-sentencepiece}, and Hugging Face Tokenizers~\citep{tokenizers} focus on efficiency and reproducibility, making them suitable for long-context, high-throughput serving.

\subsection{BPE tokenizer and its variance}
The greedy, frequency-driven merges of vanilla BPE can yield units that are poorly aligned with linguistic boundaries, such as morphology. MorphBPE~\citep{asgari2025morphbpemorphoawaretokenizerbridging} integrates morpheme-aware cues to better respect morphological boundaries. Entropy-driven pre-tokenization~\citep{hu2025entropydrivenpretokenizationbytepairencoding} further uses information-theoretic boundary indicators where naïve pre-tokenization is unreliable (e.g., Chinese). 

Additionally, standard BPE produces a deterministic segmentation given fixed merges, which limits robustness to input noise and hinders compositional generalization. This motivates stochastic BPE variants such as BPE-dropout~\citep{provilkov-etal-2020-bpe}, random-BPE~\citep{saleva-lignos-2023-changes} and StochasTok~\citep{sims2025stochastokimprovingfinegrainedsubword}. Some research also argues that the common restriction from pre-token boundaries is itself limiting, proposing BPE-based extensions that learn multi-word ``superword'' units (e.g., SuperBPE~\citep{liu2025superbpe} and BoundlessBPE~\citep{schmidt2025boundless}) to improve encoding efficiency and downstream performance. Recently, the merging direction n-gram-based algorithms can be also considered as attempt in this direction~\citep{huang2025overtokenized,cheng2026conditionalmemoryscalablelookup}.

\subsection{Tokenizer in LLM}

Tokenizer design is a fundamental modeling choice and defines the atomic symbols the model attends to and therefore shapes 1) compression (hence effective context budget and compute), 2) the granularity of lexical and morphological generalization. It will influence generation speed, effective context size, memory usage, and downstream task performance~\citep{pmlr-v235-dagan24a}. 
There is considerable evidence supporting the idea that improper tokenizer design can lead to unexpected behaviors, especially in edge cases. Notable issues include: \textit{``tokenizer boundary problem}~\citep{the_art_of_prompt, phan2024understandingmitigatingtokenizationbias}, where LLMs may fail to continue generation properly when the input is truncated mid-token, and the \textit{``undertrained tokens jailbreak''} issue where some tokens have never been trained or under-trained trigger unsafe or jailbreak behavior~\citep{li2024glitchtokenslargelanguage, land2024fishingmagikarpautomaticallydetecting}. These observations suggest that it remains necessary to re-examine several widely adopted tokenizer design choices. A similar issue about the ``intermediate merging residues'' is recognized by \citet{lian2024scaffoldbpeenhancingbytepair}. However, they focus on solving the problem during training a new tokenizer, while our algorithm is much easier-to-use and does not need to retrain either the tokenizer or the model.

\section{Identifying Intermediate Merge Residues}

        

\subsection{Motivation}
In this section, we will try to answer the two questions: 
\begin{quote}
\textit{What characteristics should intermediate merge residues have? Why should they be removed from the vocabulary?}
\end{quote}


Intuitively, intermediate tokens are those tokens that initially have high frequency during BPE learning but have low frequency in the final tokenized corpus. This suggests that these tokens are used only as the \textit{``scaffold''} or \textit{``intermediate residues''}. For example, tokens \textit{``␣includ'', ``␣delet'', ``␣tremend''}\footnote{In this paper, we use the ␣ to represent a whitespace character in a token.} in the tokenizers of the Qwen family~\citep{bai2023qwentechnicalreport,qwen25,yang2025qwen3technicalreport} are intuitively a temporary state of a final token \textit{``␣include'', ``␣delete'', ``␣tremendous''} (or some similar variances). Once the complete tokens such as \textit{``␣include'', ``␣including'', ``␣includes''} and so on have been added into the vocabulary, the token \textit{``␣includ''} may have fewer occurrences and become intermediate merge residues.
Their existence is redundant, as their downstream merges should have already fulfilled their role. 
Removing them will reduce the vocabulary size and retain the actual effective part of the vocabulary. Some rare occurrences suggest some unexpected and edge cases (like misspelling or boundary issues due to special punctuation, see Section~\ref{ssec:robust}), where the low-frequency occurrence will cause insufficient learning, thereby affecting the model's performance and robustness. In summary, we need an algorithm to identify these tokens where the ratio of the final frequency is much lower than the frequency during training and remove them from the tokenizer vocabulary. Additionally, we also need the \textit{``lite''} tokenizer to be capable of handling the rare occurrences of these tokens to make sure correct and even better language performance in some edge cases.

\subsection{Identifying Algorithm}
In this section, we discuss how to identify these intermediate merge residues from pretrained tokenizers. Though we do not retrain the tokenizer or alter its basic structure, we require a corpus to estimate the token frequency. The corpus can be the original training corpus of the tokenizer, a sample from the training corpus of the models, or any general corpus whose distribution closely matches that of the real training or testing situations.

\subsubsection{Final/Intermediate Frequency Ratio}

We can identify these tokens which are ``first with high frequency but eventually merged into other tokens, becoming infrequent'' by using the ``\textit{final/intermediate frequency ratio}'' (FI ratio). Given a BPE tokenizer's merge path and a corpus, the FI ratio of a token can be expressed by the following ratio $R=F_1/(F_1+F_2)$. Here $F_1$ represents the final occurrences of the token in the entire corpus, reflecting the actual frequency of the token once the complete tokenizer has been learned and fixed. $F_2$ represents the total occurrences of all tokens that result from merging this token in its merge tree. (Note that for TikToken tokenizers, $F_2$'s calculation is slightly different,see appendix section~\ref{sec:TikToken}.) Thus, the denominator $F_1+F_2$ captures the total number of times this token appeared during the merge process when it was added to the vocabulary. Therefore, this ratio can be used to measure the frequency of a token contained in subsequent merges. If this ratio becomes too low, it indicates that the token appears infrequently by itself, making it a candidate for being an intermediate merge residue and being removed.


\begin{table}[thbp]
    \centering
    \caption{Intermediate merging residues and other low FI ratio tokens in Qwen tokenizer.}
    \label{tab:scrap_examples}
    \begin{tabular}{lc}
        \toprule
        \textbf{Token category} & \textbf{Examples}\\
        \midrule
        Intermediate merging residues & ␣includ, delet, ␣schem, ␣entert, ␣udents, elocity \\
        Short Individual Tokens & entity, eval, import, ␣incent, ␣heck, math, ␣slight, late\\
        Root, Prefixes and Suffixes & re, ere, ␣th, ent, ther, le, on, ing, ous \\
        \bottomrule
    \end{tabular}
\end{table}

However, this naive algorithm will encounter two important exceptions, as shown in Table~\ref{tab:scrap_examples}. The first consists of word roots and affixes that encode specific semantic or grammatical structures, as well as particular inflectional variants. The second includes individual words that, while complete in themselves, more frequently serve as roots for other words, such as \textit{``perform''} in \textit{``performance''} and \textit{``performing''}. Instead of intermediate residues, they are either shared by many tokens or can stand alone with a unique meaning. These tokens are more common than the typical intermediate tokens and have their own linguistic significance.

\subsubsection{Neighbor Entropy Filtering}
The key distinction between intermediate merge residues and the other two types is about the adjacent tokens. 
For the intermediate merge residues, they are intuitively incomplete parts of other words so that the context of the tokens are somehow fixed. For example, the token ``includ'' is generally expected to be followed by one token in the small set \textit{``e''}, \textit{``ing''}, \textit{``es''}, \textit{``ed''}, and so on. However, it is not the case for the other two types. For the root, they often represent a high-level and abstract concept and can occur in many words. For example, the ``re'' can be a prefix in many words like \textit{``rewrite''}, \textit{``recover''}, \textit{``refill''}, \textit{``refund''}. Similarly, tokens which can occur individually, such as ``import'' can be followed by a wide variety of words such as various Python library names.
In summary, for the two exception types, their occurrences are more flexible to their neighboring tokens, unlike those intermediate tokens, which tend to follow more fixed combinations.

Because entropy measures the uncertainty in a system, we calculate the entropy of each token's left and right neighboring tokens in the tokenized corpus. By counting the frequencies of these neighboring tokens, we can assess the entropy of the left and right neighbors for each token.
\begin{equation}
\begin{aligned}
    S_{\text{left}}(t) &= -\sum_{s\in V}p\left(\left(s,t\right)| t\right)\log p\left(\left(s,t\right)| t\right), \\
    S_{\text{right}}(t) &= -\sum_{s\in V}p\left(\left(t,s\right)| t\right)\log p\left(\left(t,s\right)| t\right),
\end{aligned}
\end{equation}
where $t$ is the target token, $V$ is the vocabulary, $p((s,t)| t)$ (or $p((t,s)| t)$) is the conditional probability on the token pair space where the right (or left) token is fixed as $t$.

The entropy describes the diversity of its neighbors. As discussed above, the two types of exception have higher entropy at both sides, while the actual intermediate merge residues will have lower entropy at least on one side. We define the \textit{entropy score} as follows:
\begin{equation}
    S(t)=\min \left\{ S_{\text{left}}(t), S_{\text{right}}(t) \right\}
\end{equation}
Finally, the identified Intermediate Merge Residue set $\mathit{IMR}$ is the set
\begin{equation}
    \mathit{IMR}=\left\{t\in V | S(t)\leq s \text{ and } R(t)\leq r\right\},
\end{equation}
where the $s$ and $r$ are hyperparameters.




\section{Removing Intermediate Merge Residues}

After identifying intermediate merge residues, we next describe how to remove them from the vocabulary without modifying other tokens in the tokenizer or finetuning the corresponding models. It requires modifying the tokenizer's encoding process to handle the rare occurrences of these tokens while preserving performance and improving robustness.

\subsection{Split}
Given an input sentence, it should first be tokenized by the original tokenizer, where some output tokens may belong to the $\textit{IMR}$ set. For these tokens, we recursively reverse the merge operations that created them (i.e., \textbf{split} these tokens into their predecessor tokens) until all resulting tokens are no longer classified as intermediate. Notice that our algorithm will never alter the \textit{``basic''} unit in vocabulary. This step will not lead to unknown token issues, as it can at least be reverted to the original basic letters. The step also preserves the structure learned during BPE training, only \textit{``masking''} some intermediate tokens in the vocabulary.


\subsection{Re-Merge}
The splitting process may sometimes lead to excessively fragmented tokens, increasing sequence length and inefficiency.
To mitigate this effect, we apply a \textbf{re-merge} step where some fragmented tokens are re-merged into a single token, based on the \textit{``unmasked''} learned merges. See Table~\ref{tab:remerge_example} for two examples. Here, three tokens are identified as intermediate merging residues: \textit{``ruptions''}, `\textit{`\visiblespace democr''} and \textit{``atically''} which are split into basic elements like \textit{``atic''} or individual letters like \textit{``r''} and \textit{``s''}. After the re-merge step, these two words are re-organized into two more compact tokens, with clearer semantics and better linguistic alignment, such as the complete word \textit{``\visiblespace corruption''} and the plural suffix \textit{``s''}. Here, tokens in the $\textit{IMR}$ set can be seen as an indication of ``suboptimal tokenization'', suggesting the sentence meets the corner case where an infrequent and incomplete token occurs. Because our \textit{split} step introduces a chance to split these suboptimal merges, these better combinations can \textbf{re-merge} into tokens that are more aligned with linguistic structure.
In summary, this cleanup operation reduces unnecessary fragmentation and restores compact representations where appropriate, without reintroducing the removed intermediate tokens. 

\setlength{\tabcolsep}{3pt}

\begin{table}[t]
    \centering
    \caption{\textbf{Examples of the split and re-merge.} Identified intermediate merge residues are labeled by \textit{italic}. Re-merge shows a similar number of tokens and better linguistic alignment without intermediate merging residues.}
    \label{tab:remerge_example}
    \small
    \begin{tabular}{ccc}
    \toprule 
        \textbf{Original} & \textbf{Split} & \textbf{Re-merged} \\ \midrule
        \visiblespace cor\toksep\textit{ruptions} & \visiblespace cor\toksep ru\toksep ption\toksep s & \visiblespace corruption\toksep s \\
        \visiblespace \textit{democr}\toksep\textit{atically} & \visiblespace dem\toksep oc\toksep r\toksep atic\toksep ally & \visiblespace democratic\toksep ally \\
        \bottomrule
    \end{tabular}
\end{table}


\subsection{Output}
Above, we discuss how to tokenize a sentence using our LiteToken. To remove these tokens, we also need to mask them during the LLM's output/generating process. Notice that the output layer in an LLM is a linear mapping from hidden dimension to the output vocabulary size, represented as a matrix $O\in\mathbb{R}^{d_h\times N_{\text{vocab}}}$. Each column corresponds to a key vector for the corresponding token, allowing us to directly remove the columns to prevent the output/generation of the token. During the autoregressive generation process, we do not use the re-merge techniques, as re-merging may alter the previously generated tokens and break the KV cache. Compatibility with the KV cache and re-merging is left as a direction for future work.

\section{Experiments}

In this section, we design experiments to demonstrate the prevalence of intermediate tokens in modern tokenizers and the vulnerability of these tokenizers to these tokens. Then, our LiteToken can help model to save parameters, improve the robustness while maintaining overall performance.

\subsection{Tokenizers, Languages, and Datasets}
We study the tokenizers of six state-of-the-art models: the o200k\_base tokenizer from GPT family~\citep{gpt4o} (models after GPT-4o and GPT-oss), the Qwen tokenizer~\citep{bai2023qwentechnicalreport,qwen25,yang2025qwen3technicalreport} used by all Qwen models, the newest Deepseek tokenizer used by models after Deepseek-V3~\citep{deepseekv3},  BLOOM~\citep{bloom}, Llama-3~\citep{llama3} and Gemma-3~\citep{gemma3}. Notice that Llama-3, Gemma-3 and GPT use the modified version of the BPE algorithm provided by TikToken~\citep{TikToken}, marked by the $^*$.


In this paper, we focus on the Latin tokens (primarily in English), which are fully composed of ASCII characters. Since the exact pretraining corpora of modern LLMs are not fully disclosed and are often infeasible to reconstruct in a reproducible way, we use C4-en~\citep{c4} as a large-scale, broad-coverage English web corpus that serves as an estimation for the statistics encountered in a general-purpose LLM setting. With approximately 171 billion tokens (when tokenized by the original Qwen tokenizer), it covers a wide range of natural language cases, making it well-suited for analyzing token usage patterns and approximating the data distribution under which modern tokenizers are typically trained and applied. While similar behaviors can also be observed in other languages and domains, they are not the primary focus of this study and are therefore presented in the appendix.

\begin{table*}[htpb]
    \centering
     \caption{\textbf{Prevalence of intermediate tokens across tokenizers.} We report the number and proportion of tokens identified as intermediate before and after applying the neighbor-entropy filter (denoted as \emph{Ent.}). We select a moderate value as our filter: tokens with FI ratio$<0.25$ and entropy score $<4$; and for the TikToken tokenizers$^*$, tokens with FI ratio$<0.05$ and entropy score $<3.5$ }
    \label{tab:scrap_stats}
    \resizebox{\textwidth}{!}{
     \small
    \begin{tabular}{lccccc}
        \toprule
        \textbf{Model}& \textbf{Vocab \#(en)} & \textbf{Scrap \# (w/o Ent.)} & \textbf{Scrap \# (Ent.)} & \textbf{Scrap \% (w/o Ent.)} & \textbf{Scrap \% (Ent.)} \\
        \midrule
        Qwen   & 94376  &  7216 &  4784 & 7.65   &  5.07 \\
        DeepSeek-V3   & 72281  &  7037 &  4686 & 9.74   &  6.48 \\
        BLOOM  & 107963  &  6441 &  4394 & 5.97   &  4.07 \\ \midrule
        Gemma-3$^*$  & 150426  &  23510 &  15584 & 15.63   &  10.36 \\
        Llama-3$^*$  & 98102  &  12737 &  7966 & 12.98   &  8.12 \\
        GPT-o200k\_base$^*$   & 128856  &  17340 &  10337 & 13.46   &  8.02 \\
        \bottomrule
    \end{tabular}
    }
\end{table*}

\subsection{Prevalence of Intermediate Tokens}



Using token frequencies computed on C4-en, we report the number and proportion of tokens identified as intermediate under a pure ratio filter and after applying the neighbor-entropy filter. Considering the different structures for the standard BPE and TikToken ones, we use different thresholds for these two types. For the standard BPE tokenizers (Qwen, DeepSeek and BLOOM), we choose the FI ratio threshold as $0.25$ and the entropy score threshold as $4.0$. The tokenizer from TikToken (Llama-3, GPT and Gemma-3) use FI ratio threshold of $0.05$ and an entropy score threshold of $3.5$. We will discuss the choice of the hyperparameters in appendix section~\ref{ssec:hyper_param}.

\begin{figure}[t]
    \centering
    \includegraphics[width=0.45\linewidth]{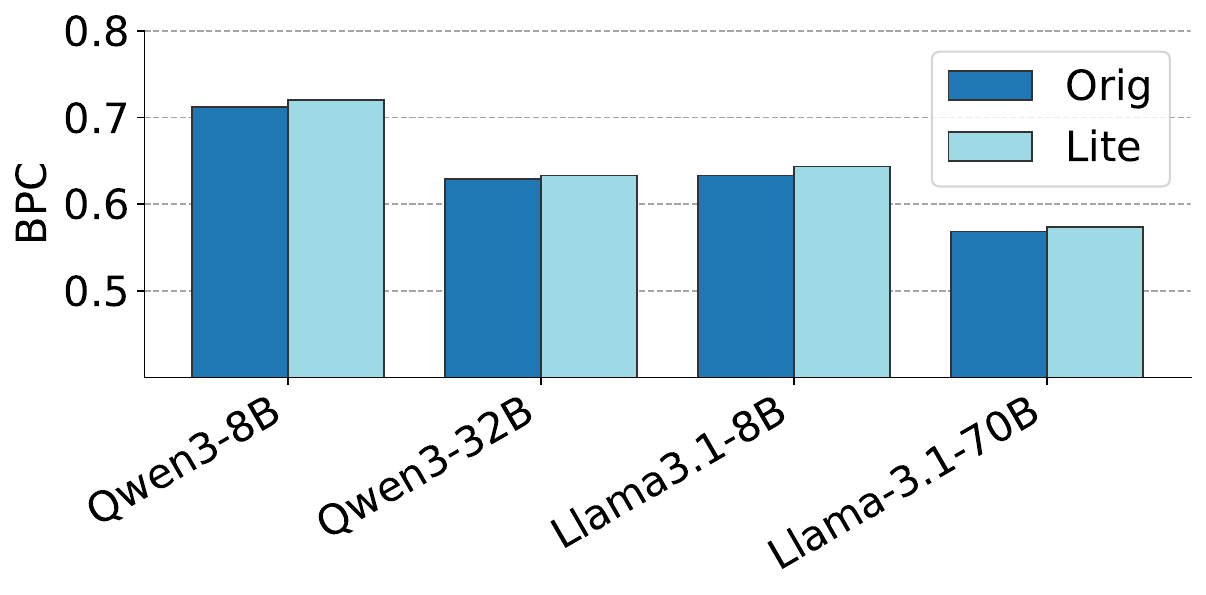}
    \includegraphics[width=0.45\linewidth]{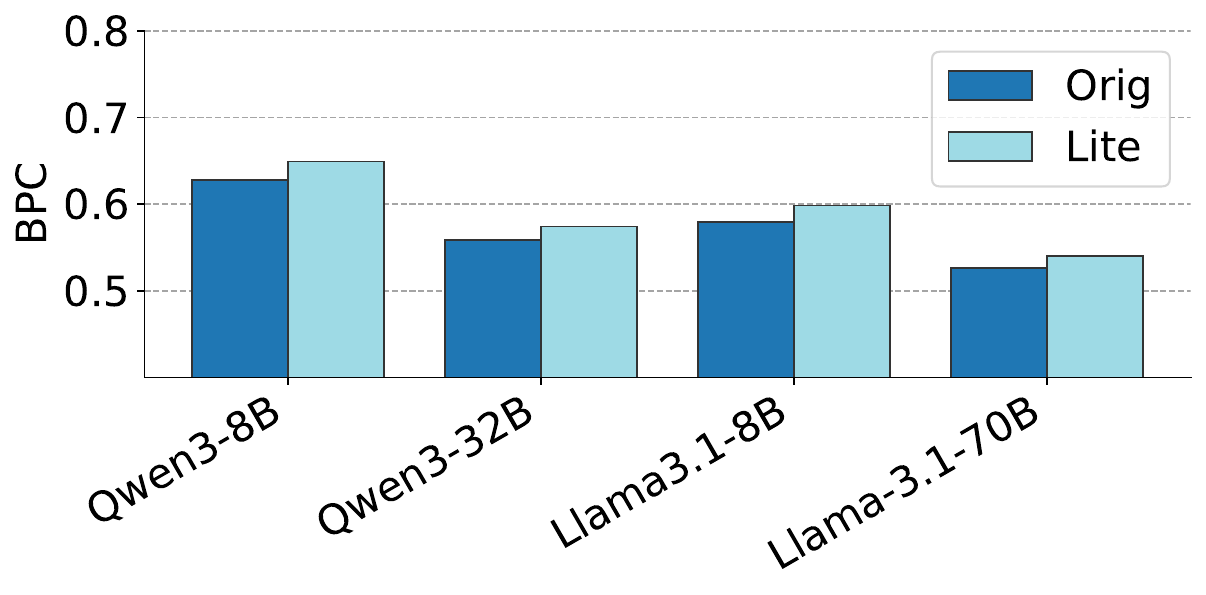}
    \caption{\textbf{Bits-per-character (BPC) on sampled sentences with intermediate tokens.} \textbf{Left}: C4-en; \textbf{Right}: RedPajama}
    \label{fig:ppl}
\end{figure}

The results are summarized in Table~\ref{tab:scrap_stats}. For all of these six models, about $5\%$ to $10\%$ tokens are identified by our LiteToken. 
Furthermore, considering the N-grams like DeepEncoding~\citep{huang2025overtokenized} or Engram~\citep{cheng2026conditionalmemoryscalablelookup}, the effectiveness of pruning will be enlarged exponentially. With the DeepSeek-V3 tokenizer, our algorithm can help the 3-gram vocabulary to shrink about 22\%, largely reduce the memory usage and avoid hash conflict. 
This observation motivates our subsequent experiments, which evaluate whether removing such intermediate tokens affects generation quality and robustness. The result of other languages and domains can be seen in appendix section~\ref{sec:multi}.

\subsection{Language Modeling}
\label{ssec:exp_lm}

Language modeling is the fundamental capacity of an LLM. In this experiment, we evaluate how removing intermediate merging residues affects the language modeling performance. Here, we choose Qwen3-8B, Qwen3-32B, Llama-3.1-8B and Llama-3-70B as tested models, where we do not finetune the model but only modify the tokenizer.
We evaluate language modeling capacity on a sampled C4-en dataset. As a supplement, we also evaluated the arXiv and Wikipedia subsets of RedPajama-1T~\citep{weber2024RedPajama} as data sources that are relatively orthogonal to C4’s web-based corpus, in order to assess the corpus-level generalization of our results. 
We sampled 10,000 sentences from these two corpora that contain at least one intermediate token (since our algorithm makes no modifications to sentences that do not include any intermediate-merge tokens).

We report bits-per-character (BPC) and average tokenized sequence length in Figure~\ref{fig:ppl} and Table~\ref{tab:length}. Removing intermediate merging residues from the tokenizer affects BPC and the number of tokens in a negligible way, suggesting that all models can directly use this pruned tokenizer without finetuning.
The difference in BPC between the original tokenizer and the LiteToken tokenizer is less than 0.2 and the difference in token numbers is also negligible (about $3\%$). This supports the conclusion that removing these tokens is almost lossless and that our LiteToken is ``plug-and-play''.

\begin{table}[t]
    \centering
    \caption{\textbf{Average token numbers.} Models in one family use the same tokenizer, having the same number of tokens.}
    \label{tab:length}
    \small
    \begin{tabular}{lcccc}
    \toprule
        \multirow{2}{*}{\# Tokens}\hspace{3pt} & \multicolumn{2}{c}{Qwen}\hspace{3pt} & \multicolumn{2}{c}{Llama-3.1} \\ \cmidrule(lr){2-3} \cmidrule(lr){4-5}
               & Original & Lite & Original & Lite \\ \midrule
        C4-en &  171.16 & 173.76 & 171.04 & 173.71\\ 
        RedPajama &    199.54  & 204.30 & 199.77 & 208.70 \\
        \bottomrule
    \end{tabular}
\end{table}

\subsection{Rare Token Robustness}
\label{ssec:robust}

In the introduction, we noted that rare tokens, such as those caused by misspellings or adversarial perturbations, can expose models to under-training effects, making them vulnerable to attacks and resulting in unstable generation. In this section, we examine whether our method can effectively enhance robustness under such conditions. 



\paragraph{Setup.}
Here, we construct a test set where sentences contain ``misspelling'' or ``rare input'', achieved by introducing single-token perturbations into otherwise clean English inputs. Each perturbation targets a token identified as intermediate according to our criteria. As an example, a word \textit{``assembly''} could be perturbed to \textit{``asssembly''} and then the correct tokenization \textit{``assembly''} (one token) could be changed to \textit{``ass''} and \textit{``sembly''}, where the \textit{``sembly''} is a rare token and may be confusing for some models. Or some cases where the punctuation or special symbols create unnatural token boundaries (e.g., \textit{.assembly} $\to$ \textit{.as} $+$ \textit{sembly}).
Our experiments aim to answer whether the misspelling or unnatural inputs lead to performance degradation, and whether our LiteToken can address this issue. We construct 10,000 examples, which are all extracted from the ``question'' part of the HotpotQA dataset\citep{hotpotqa}. 

Here we test two capacities of the models:
\begin{itemize}
    \item \textbf{Explicit perturbation identifying and correctness}: We asked the models to explicitly identify the perturbed words and provide the corrected ones. This tests whether the models can correctly identify unusual tokens and understand the appropriate context in which they should be interpreted. We measure the accuracy of identifying and correcting the perturbation for each sentence simultaneously.
    \item \textbf{Perturbed question answering}: The models are required to answer the question with perturbations present. This tests whether the model can still understand and answer the original question correctly, even under perturbations. 
\end{itemize}
The results are shown in Table~\ref{tab:perturb}. Models equipped with LiteToken achieve higher accuracy on both error correction and perturbed question answering. In contrast, models without LiteToken exhibit clear degradation under perturbations (1.78\% and 0.98\% for Qwen-3-8B and Qwen-3-32B, respectively), whereas LiteToken reduces the drop to 0.82\% and 0.29\%.
\begin{table}[t]
    \centering
    \caption{\textbf{Model robustness under perturbation in Error Correction(EC) and Question Answering(QA) with bootstrap variance.} The ``-Lite'' is with our LiteToken. * is the result of unperturbed questions.}
    \label{tab:perturb}
    \small
    \begin{tabular}{lcccc}
    \toprule
       \textbf{Model}  & \textbf{Acc.(EC)} \% & \textbf{F1(QA)} \%  & \textbf{F1(QA*)} \% & \textbf{$\Delta$ F1} \% \\ \midrule
       Qwen-3-8B  & 89.77$\pm$0.01 & 79.66$\pm$0.03  & \textbf{81.44}$\pm$0.01 & 1.78\\
       Qwen-3-8B-Lite  & \textbf{90.69}$\pm$0.01 & \textbf{80.16}$\pm$0.03 & 80.98$\pm$0.01 & \textbf{0.82}\\ \midrule
       Qwen-3-32B  & 92.25$\pm$0.01 & 83.12$\pm$0.01 & \textbf{84.10}$\pm$0.01 & 0.98\\
       Qwen-3-32B-Lite & \textbf{93.84}$\pm$0.01 & \textbf{83.30}$\pm$0.01 & 83.59$\pm$0.01 & \textbf{0.29}\\ \midrule
    \end{tabular}
\end{table}

\subsection{Downstream Tasks}

To verify that intermediate merge residue removal does not affect standard NLP performance, we evaluate our method on multiple benchmarks, including SQuAD~\citep{squad}, GSM8K~\citep{gsm8k}, HellaSwag~\citep{hellaswag}, ARC~\citep{arc}, MMLU~\citep{mmlu} and GPQA~\citep{gpqa}, using Qwen-3 series models and Llama-3.1-8B. We compare the original tokenizers with our LiteToken under identical model parameters and decoding settings, with model temperature set to 0(for thinking tasks, the temperature is 0.6). Following standard practice, we report the F1 and accuracy scores. Moreover, we also report the percentage of samples that have at least one intermediate token(note as w/change), ensuring that LiteToken actually changed the tokenization inside many of the benchmark prompts. The results are shown in Table~\ref{tab:gqa}. Across those models and tasks, we observe that removing intermediate tokens can still maintain performance on question answering, reasoning, math abilities, etc.

\begin{table*}[t]
    \centering
    \caption{\textbf{Model performance on downstream tasks.} All tasks are done 0-shot. ARC uses the Challenge division. Chain of Thought(CoT) is only enabled in GSM8K and GPQA tasks. We also reported the percentage of samples that have at least one intermediate token(note as w/change). }
    \label{tab:gqa}
    \scriptsize
    \begin{tabular}{lcccccccccccc}
    \toprule
        \multirow{2}{*}{Metric}\hspace{3pt} & \multicolumn{3}{c}{Llama-3.1-8B}\hspace{3pt} & \multicolumn{3}{c}{Qwen-3-8B} & \multicolumn{2}{c}{Qwen-3-4B} & \multicolumn{2}{c}{Qwen-3-1.7B} & \multicolumn{2}{c}{Qwen-3-0.6B}\\ \cmidrule(lr){2-4} \cmidrule(lr){5-7} \cmidrule(lr){8-9} \cmidrule(lr){10-11} \cmidrule(lr){12-13}
               & Original & Lite & w/Change & Original & Lite & w/Change & Original & Lite & Original & Lite & Original & Lite \\ \midrule
        SQuAD(F1)&    0.806  &   0.804  &   19.89 &   0.833  &  0.829 &   16.74 &   0.797  &  0.793 &   0.743  &  0.735 &   0.652  &  0.641  \\ 
        GSM8K(CoT)&    84.53  &   84.38  &   73.95 &   93.71  &  93.56 &   80.64 & 89.23  &   89.08  &   77.86  &  77.56 & 57.24  &   57.16\\
        HellaSwag &    70.14  &   69.36   &   32.43 &   67.32  &  67.06 &   74.53 &  62.23  &   62.01   &   54.37  &  54.25  & 48.74  &   48.56\\ 
        ARC-Challenge &    56.23  &   56.14   &   33.69 &   66.81  &  66.72 &   50.31 &   60.32  &  60.24 &   54.10  &  54.01 &   46.84  &  46.84  \\
        MMLU&    68.39  &   68.12   &   25.25 &   78.35  &  78.21 & 73.07 &   39.38  &  72.98 &  65.93  & 65.87    & 59.32   & 59.26 \\ 
        GPQA(CoT) &    42.93  &   42.42   &   31.31 &   44.44  &  43.94 &   31.31 &  37.88  &  37.37 &   30.30  &  29.80   &  27.78  &  27.78\\
        \bottomrule
    \end{tabular}
\end{table*}


Taken together, the robustness experiments and downstream evaluation indicate that intermediate tokens primarily affect stability under noisy conditions, while having a limited impact on performance in standard evaluation settings.

\section{Conclusion}

This paper studies intermediate merge residues, a previously overlooked artifact of the original BPE training process. Through a comparative analysis of modern tokenizers, we show that such residues are widely present across different tokenizer families. We then propose LiteToken, which removes these redundant tokens to reduce token-related parameters, lower memory and computation cost, and improve linguistic alignment. Experiments on Qwen-3 and Llama-3.1 demonstrate that LiteToken reduces excessive fragmentation during generation and improves robustness under challenging tokenization conditions such as misspellings and boundary artifacts, while preserving language modeling and question-answering performance in standard settings. These results highlight the importance of tokenizer design and suggest that intermediate merge residues deserve closer attention in future NLP systems.

\section{Limitations}
Our work has several limitations. First, our study focuses on English, and the behavior of intermediate merge residues in morphologically rich (like Arabic or Finnish) or non-alphabetic languages (like Chinese) remains largely unexplored.
Second, although we target a ``plug-and-play'' setting, our approach operates purely at the tokenization level and does not account for potential interactions with model retraining or adaptation, which could further improve the robustness and performance. We leave these directions for future work. Another limitation is that, for different tokenizers, we use the same datasets to estimate the frequencies. The best method is to use the actual training corpus os these tokenizers, but it is not fully open-sourced and available.


\bibliographystyle{iclr2027_conference}
\bibliography{custom}

\newpage

\appendix

\section{For TikToken Tokenizers}
\label{sec:TikToken}
Although our algorithm works with standard BPE tokenizers, some prevailing tokenizers, such as those trained with TikToken~\citep{TikToken} (e.g., llama-3 and GPT families), use a modified BPE algorithm.
Unlike a standard BPE tokenizer, whose encoding process strictly replays the training process, by sequentially applying merges in the learned order, a TikToken-trained tokenizer does not record a strict merge tree. Instead, it only stores the resulting token from each merge, without tracking which two tokens were merged to create it.

During encoding, because the strict merge tree is absent, the tokenizer permits all possible merge that yield one same token. For example, when the encoding a token like \texttt{abcd}, suppose it was originally learned by merging \texttt{ab} and \texttt{cd} during training. In a standard BPE tokenizer, only the merge ``\texttt{ab} $+$ \texttt{cd} $\to$ \texttt{abcd}'' would be allowed; in TikToken, however, if \texttt{a} and \texttt{bcd} are also higher-ranked tokens, then ``\texttt{a} $+$ \texttt{bcd} $\to$ \texttt{abcd}'' is also allowed. As a result, each token in a TikToken-trained tokenizer can be viewed as having multiple potential parent pairs.

Given the widespread use of such tokenizers, our algorithm is designed to be compatible with them. Specifically, for this setting, we consider all permitted merges and perform frequency statistics using the merge table actually used during encoding, ensuring alignment with the model’s real usage. In the split step, we revert each occurrence of a token back to the particular decomposition from which it was merged. As a result, TikToken tokenizers has different distributions on those intermediate tokens, which is mentioned in the following ablation studies.

\section{Ablation Studies}

\label{ssec:hyper_param}
 \subsection{Threshold} 
\begin{figure*}[htpb]
     \centering
     \includegraphics[width=0.45\linewidth]{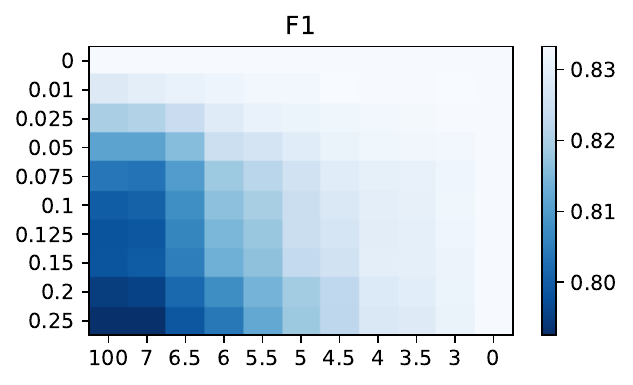}
     \includegraphics[width=0.45\linewidth]{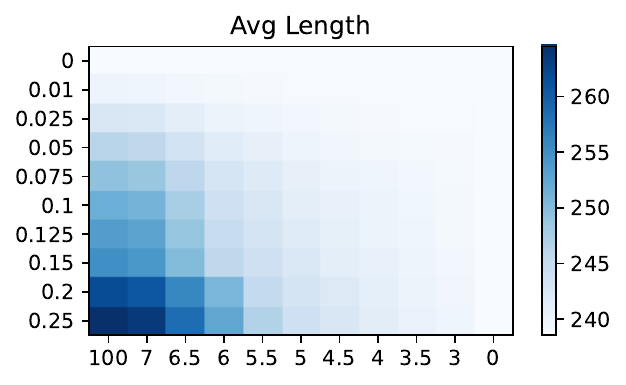}
     \\
     \includegraphics[width=0.45\linewidth]{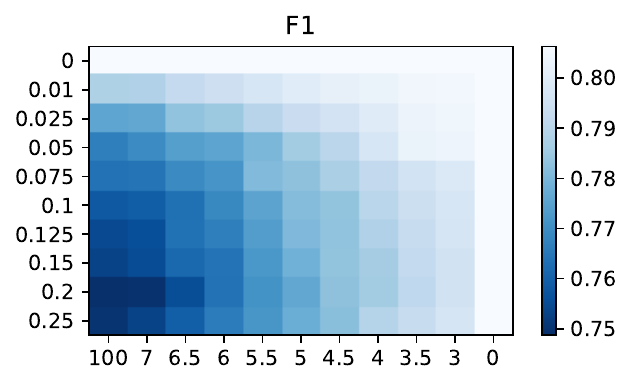}
     \includegraphics[width=0.45\linewidth]{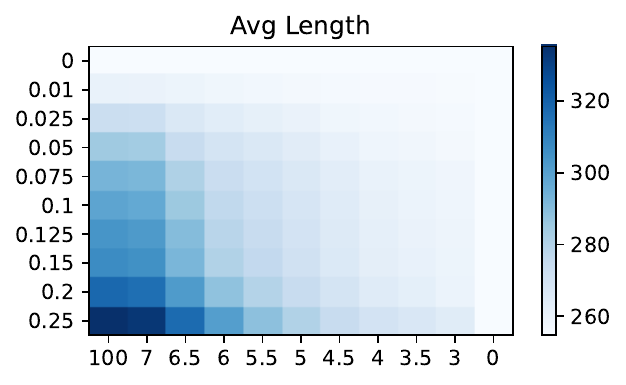}
     \caption{\textbf{Ablation study of the threshold of frequency ratio and entropy.} The columns are entropy, and the rows are ratio. \textbf{Above}: Qwen; \textbf{Below}: Llama}
     \label{fig:ablation}
 \end{figure*}

Here we operate an ablation study of two important hyper-parameters: the threshold of frequency ratio and entropy. Here, Qwen-3-8B and Llama-3.1-8B are selected as representations for standard BPE and TikToken BPE respectively. Shown in Figure~\ref{fig:ablation}, the results indicate that within our setting ratio $0.25$ and entropy $4$ for standard BPE (above) and ratio $0.05$ and entropy $3.5$ for TikToken BPE (below), the f1 score (SQuAD) and the count of tokens do not show significant influence so we choose them. 

\subsection{Split \& Re-merge.}
Here we focus on the effect of the two steps ``\textbf{split}'' and ``\textbf{re-merge}''. The results are shown in Table~\ref{tab:ablation_split_remger}. The split step remove about 5,000 tokens from the vocabulary but lightly influence the model's behavior, while the re-merge step can reduce the PPL and the number of tokens and recover the F1 score on SQuAD, suggesting the effectiveness of the ``re-merge'' step.

  \begin{table}[htpb]
    \centering
    \caption{\textbf{Ablation on the two steps: split and re-merge.} Using Qwen-3-8B as test model. BPC and token length are calculated on c4 dataset, while F1 is from SQuAD.}
    \label{tab:ablation_split_remger}
    \small
    \begin{tabular}{lcccc}
    \toprule
        Tokenizer & \# vocab & BPC & \# tokens  & F1 \\ \midrule
        Original tokenizer & 94376   & 0.708  &   171.16  &  0.8329   \\
        LiteToken w/o remerge & 89592  & 0.770 & 174.51 & 0.8288 \\
        LiteToken & 89592 & 0.717  & 173.76 & 0.8292 \\ \bottomrule
    \end{tabular}
\end{table}

\section{Prevalence of Intermediate Tokens in Multiple Domains}
\label{sec:multi}

\subsection{Pure English Tokenizer}

Though the data used to filter the tokenizers is an English dataset (C4-en), the tokenizers being filtered are still multilingual. Thus, it is not clear if the tokens being removed are simply non-English tokens. So we reported the intermediate tokens of GPT-NeoX tokenizer~\citep{black2022gptneox20bopensourceautoregressivelanguage}, which is a pure BPE English tokenizer. The result is shown in Table~\ref{tab:scrap_stats_multi}, which is close to that of Qwen, indicating the validity of the multilingual tokenizers' data.

\subsection{Intermediate Tokens in Code and Other Languages}

\begin{CJK*}{UTF8}{gbsn}
In this paper, we focus on the Latin tokens (primarily in English), which are fully composed of ASCII characters. In our preliminary experiments, we also found similar issues in other languages, though each language also have its own characteristics. For example, some Chinese four-character idioms in the vocabulary have the corresponding three-character intermediate tokens, which can never be used independently in Chinese. For example, in Qwen tokenizer, ``莫名其妙''(id: 115135), meaning ``confusing'' in Chinese, is merged from a meaningless intermediate token ``莫名其''(id: 115134) and ``妙''(id: 100279). 
\end{CJK*}

Here, we use C4's French(fr), German(de), Japanese(ja), Arabic(ar) and Chinese(zh,simplified) subset as monolingual dataset, the Vault~\citep{manh2023vaultcomprehensivemultilingualdataset} as code dataset, and Wikipedia as multilingual dataset. The Vault contains 10 programming languages, while Wikipedia contains 323 different languages, suitable for mixture-of-language test. Then, we run the same LiteToken pipeline as English experiments. The results are summarized in Table 9. 

For French and German, since we count all tokens that only have ASCII and their special characters, many of the tokens overlap with English and are seldom used, resulting in their scrap token percentage lower than English. Japanese, Arabic and Chinese all have their independent vocabulary(note that Japanese kanji and simplified Chinese characters are different), they all have their independent results, which are quite small compared to Latin letter languages.

As for the mixtures, ...

\begin{table*}[htpb]
    \centering
    \caption{\textbf{Prevalence of intermediate tokens across code and other languages.} We report the number and proportion of tokens identified as intermediate before and after applying the neighbor-entropy filter (denoted as \emph{Ent.}). We select a moderate value as our filter: tokens with FI ratio$<0.25$ and entropy score $<4$; and for GPT-o200k\_base$^*$(TikToken), tokens with FI ratio$<0.05$ and entropy score $<3.5$ ;}
    \label{tab:scrap_stats_multi}
     \small
    \begin{tabular}{lccccc}
        \toprule
        \textbf{Lang/Domain} & \textbf{Vocab \#(en)} & \textbf{Scrap \# (w/o Ent.)} & \textbf{Scrap \# (Ent.)} & \textbf{Scrap \% (w/o Ent.)} & \textbf{Scrap \% (Ent.)} \\
        \midrule
        \multicolumn{6}{c}{\textit{Qwen}} \\
        \midrule
        en & 128856  &  17340 &  10337 & 13.46   &  8.02 \\
        fr & 134482  &  14777 &  8493  & 10.99   &  6.32 \\
        de & 136947  &  13561 &  6994  & 9.90   &  5.11 \\
        ja & 6452  &  29 &  28  & 0.45   &  0.43 \\
        ar & 2714  &  94 &  28  & 3.46   &  1.03 \\
        zh & 24875 & 528 &  479  & 2.11   & 1.93 \\
        multilingual & 149999  &  6548 &  5407  & 4.37   &  3.60 \\
        code & 128856  &  11473 &  8463  & 8.90   &  6.57 \\
        \midrule
        \multicolumn{6}{c}{\textit{GPT-o200k\_base$^*$}} \\
        \midrule
        en & 94376  &  7216 &  4784 & 7.65   &  5.07 \\
        fr & 96160  &  3680 &  2981  & 3.83   &  3.10 \\
        de & 97039  &  3524 &  2797  & 3.63   &  2.88 \\
        ja & 26927  &  400 &  364  & 1.49   &  1.35 \\ 
        ar & 2291  &  179 &  174  & 7.81   &  7.59 \\
        zh & 7958  &  115 &  106  & 1.45   &   1.33 \\
        multilingual & 199985  &  13176 &  8825 & 6.59   &  4.41 \\
        code & 94384  &  5100 &  4039  & 5.40   &  4.28 \\
        \midrule
        \multicolumn{6}{c}{\textit{GPT-NeoX}} \\
        \midrule
        en & 47679 & 5117 & 4342 & 10.72 & 9.11 \\
        \bottomrule
    \end{tabular}
\end{table*}
     
\end{document}